\documentclass[
  journal=medium,
  manuscript=article-type,
  year=2020,
  volume=37,
]{cup-journal}

\usepackage{amsmath}
\usepackage[nopatch]{microtype}
\usepackage{booktabs}
\usepackage{multirow}

\title{Constructing Vec-tionaries to Extract Message Features from Texts: A Case Study of Moral Content}

\author{Zening Duan}
\affiliation{School of Journalism and Mass Communication, University of Wisconsin-Madison, Madison, 53706, WI, United States}

\author{Anqi Shao}
\affiliation{School of Journalism and Mass Communication, University of Wisconsin-Madison, Madison, 53706, WI, United States}

\author{Yicheng Hu}
\affiliation{Department of Chemical and Biological Engineering, University of Wisconsin-Madison, Madison, 53706, WI, United States}

\author{Heysung Lee}
\affiliation{School of Journalism and Mass Communication, University of Wisconsin-Madison, Madison, 53706, WI, United States}

\author{Xining Liao}
\affiliation{School of Journalism and Mass Communication, University of Wisconsin-Madison, Madison, 53706, WI, United States}

\author{Yoo Ji Suh}
\affiliation{School of Journalism and Mass Communication, University of Wisconsin-Madison, Madison, 53706, WI, United States}

\author{Jisoo Kim}
\affiliation{School of Journalism and Mass Communication, University of Wisconsin-Madison, Madison, 53706, WI, United States}

\author{Kai-Cheng Yang}
\affiliation{Network Science Institute, Northeastern University, Boston, 02115, MA, United States}

\author{Kaiping Chen}
\affiliation{School of Journalism and Mass Communication, University of Wisconsin-Madison, Madison, 53706, WI, United States}

\author{Sijia Yang}
\affiliation{School of Journalism and Mass Communication, University of Wisconsin-Madison, Madison, 53706, WI, United States}
\email[F. Author]{syang84@wisc.edu}

\keywords{computational text analysis, message feature, moral content, word embedding, optimization, crowd-sourcing} %% First letter not capped

\begin{document}

\begin{abstract}
While researchers often study message features like moral content in text, such as party manifestos and social media, their quantification remains a challenge. Conventional human coding struggles with scalability and intercoder reliability. While dictionary-based methods are cost-effective and computationally efficient, they often lack contextual sensitivity and are limited by the vocabularies developed for the original applications. In this paper, we present an approach to construct vec-tionary measurement tools that boost validated dictionaries with word embeddings through nonlinear optimization. By harnessing semantic relationships encoded by embeddings, vec-tionaries improve the measurement of message features from text, especially those in short format, by expanding the applicability of original vocabularies to other contexts. Importantly, a vec-tionary can produce additional metrics to capture the valence and ambivalence of a message feature beyond its strength in texts. Using moral content in tweets as a case study, we illustrate the steps to construct the moral foundations vec-tionary, showcasing its ability to process texts missed by conventional dictionaries and word embedding methods and to produce measurements better aligned with crowdsourced human assessments. Furthermore, additional metrics from the vec-tionary unveiled unique insights that facilitated predicting outcomes such as message retransmission.  
\end{abstract}

*This version has not undergone peer review.* 

\section{Introduction}
\label{section:intro}
Social scientists from various disciplines have worked on improving the quantitative measurement of message features, such as emotions \autocite{brady2017emotion}, uncivil and gendered language \autocite{theocharis2016bad, chen2024gender}, and more recently, moral intuitions \autocite{graham2013moral, clifford2013words,weber2021extracting,zhou2022moral}. This exploration extends across diverse text sources, including government records, newspapers, social media posts, and other unstructured textual repositories. However, quantifying message features from texts presents a formidable challenge. For example, human coding cannot easily scale up to process “big data” \autocite{hopkins2010method}, or in some cases, is suboptimal to alternative measurement strategies such as crowdsourcing, particularly when intercoder reliabilities fall short of conventional threshold \autocite{weber2021extracting}. The rise of computational content analysis methods, notably text-as-data approaches \autocite{grimmer2022text}, has popularized the use of dictionaries as a low-cost, quick-to-use measurement strategy for handling large-scale textual data. However, this approach has inherent limitations, lacking sensitivity to context-specific applications and often encountering difficulties in extracting signals from short-format texts like tweets, due to its fixed and limited vocabulary.

This study introduces “vec-tionaries,” a novel computational method for extracting message features. We use moral content as a case study to demonstrate its advantages. Drawn from the Moral Foundations Theory (MFT) \autocite{graham2009liberals,haidt2012righteous}, moral content belongs to a category of message features that invoke and appeal to fundamental moral values. MFT has reshaped scholarly understanding of morality and how it relates to the formation of political attitudes and expression. In brief, MFT suggests that individuals’ moral intuitions are rooted in six major psychological systems, or foundations, including Care/Harm, Fairness/Cheating, Loyalty/Betrayal, Authority/Subversion, Sanctity/Degradation, and Liberty/Oppression. Each of these foundations acts like a “taste bud,” allowing individuals to quickly judge situations in the social world that uphold or violate these foundations through gut-like reactions of likes and dislikes \autocite{haidt2012righteous}. For instance, the Care/Harm foundation involves sensitivity toward the suffering of vulnerable beings, such as immigrants, while those attuned to Authority/Subversion tend to prioritize social hierarchy and tradition. 

Moral words may do “the work of politics.” In recent years, MFT has been employed not only to analyze the moral rhetoric found in party manifestos \autocite{jung2020mobilizing}, speeches \autocite{graham2009liberals}, and state legislatures \autocite{mucciaroni2011debates}, but also to understand how individuals across the political spectrum process arguments and mold behaviors. A growing body of research demonstrates that moral foundations play a crucial role in fueling partisan disagreements on environmental attitudes \autocite{feinberg2013moral}, candidate trait evaluations \autocite{clifford2014linking}, and voting choices \autocite{jung2020mobilizing}. Furthermore, studies indicate that moral content can induce attitude changes \autocite{clifford2013words,feinberg2013moral}, increase the retransmission of gender-biased videos \autocite{chen2024gender}, and predict hate speech on social media \autocite{solovev2023moralized}. Since a person’s moral values are not always visible on the surface, individuals often express their moral values through words. Morally relevant content or moral sentiment serves as a means to communicate these values to others. This emphasizes the importance of developing accessible, interpretable, and scalable computational tools for extracting moral content from textual data, which is the gap our study aims to address.

Our vec-tionaries approach leverages the semantic relations between validated dictionary words encoded in pre-trained word embeddings, where the message features can be represented as semantic axes residing in the same semantic vector space \autocite{an2018semaxis,kozlowski2019geometry}. Our model then identifies these axes through a nonlinear optimization algorithm. Users can then project unseen messages onto these axes to measure the message features of interest. Compared with the dictionary approach, which only contains the semantic information of a limited vocabulary, a vec-tionary incorporates additional signals from other words outside the original dictionary’s vocabulary by exploiting their embeddings-based semantic relations. Moreover, pre-trained word embeddings allow a vec-tionary to capture contextual information in documents and quantify additional properties of the message feature such as \textit{Valence} and \textit{Ambivalence}, without relying on human-labeled data for supervised classifier training. While our study focuses on moral content to illustrate the measurement advantages, conceptual foundations, and implementation protocols of vec-tionaries, we note that the methodology for constructing vec-tionaries extends beyond moral content and can be applied to measure various message features, such as emotions, frames, incivility, and many more.

Next, we overview strengths and weaknesses of existing computational methods for measuring moral content in Section \ref{section:existing methods}. Section \ref{section:vectionary_approach} introduces our vec-tionary approach, and three metrics derived from it to capture different aspects of moral content in texts. Section \ref{section:model_validation} compares vec-tionary to the state-of-art moral foundations dictionary using crowdsourced annotations from two million COVID-19 tweets, showing that our approach is superior to or at least on par for measuring moral content. Section \ref{section:application} applies our vec-tionary to study extract moral content from the same tweet corpus, predicting retweets and demonstrating additional value in enhancing empirical research on moral content. Section \ref{section:discussion} concludes; proofs, illustrations, and supporting information are in the Supplementary Material A-K.

\section{Existing Computational Methods to Measure Moral Content}
\label{section:existing methods}

Dictionaries and word embeddings are two of the most prominent methods to extract moral content from textual data. In this section, we provide an overview of these two measurement strategies and discuss their strengths and limitations. 

\subsection{Moral foundations dictionaries}
\label{section:existing methods_dict}
In early work, Graham et al. developed the first Moral Foundations Dictionary (MFD) by using frequencies of foundation-relevant words \autocite{graham2009liberals}, particularly synonyms and antonyms, to measure differences in moral values between liberal and conservative sermons. However, the original MFD had fewer words (on average, 32 for each moral foundation) than many other dictionaries. Frimer and colleagues introduced the MFD 2.0, a more sophisticated version of the first MFD \autocite{frimer2017moral}, by proposing a much larger set of candidate words. Subsequently, the extended Moral Foundations Dictionary (eMFD)\autocite{hopp2021extended} further expanded the list to encompass approximately 3,270 English words associated with five moral foundations with varying weights. Deviating from its ancestors, eMFD assigns each word to all five moral foundations instead of exclusively to a single moral foundation. Additionally, eMFD is constructed from text annotations generated by a group of human coders (N = 557) rather than a few trained coders. As the latest addition to MFT, the Liberty/Oppression foundation was absent in most existing dictionaries, including those mentioned above. To address this, Araque et al. introduced LibertyMFD, a foundation-specific lexicon to operationalize this moral foundation \autocite{araque2022libertymfd}. 

Word count-based method has made significant strides in the textual analysis of moral content \autocite{solovev2023moralized}, especially excels at interpretability. By employing pre-established word lists, this method provides direct insights into the contributing words that define the message feature. Nevertheless, this approach has some drawbacks. Its effectiveness largely depends on the vocabulary included in the dictionary, any omission of a word results in reduced coverage. Moreover, this method often overlooks the context in which words appear. A single word might bear different meanings based on its surrounding context, a nuance often missed, making it difficult to generalize a dictionary developed in one specific context to others. The dictionary approach often suffers from inflexibility, particularly when adapting or extending the dictionary to accommodate evolving linguistic nuances, a task that can be labor-intensive. All present notable challenges and call for improvement. As a response, Garten et al. introduced the Distributed Dictionary Representation approach (DDR) \autocite{garten2018dictionaries}, and An et al. proposed the SEMAXIS framework, both utilizing word-embedding to better quantify short-form texts from contextually dependent data \autocite{an2018semaxis}. In the next section, we provide detailed explanations of these word embedding approaches and then illustrate how our moral foundations vec-tionary is designed and implemented building upon these efforts.

\subsection{Word embeddings and distributed dictionary representations}
\label{section:existing methods_embed}
In the field of natural language processing, significant progress has been made in learning effective representations of words as vectors in high-dimensional semantic spaces \autocite{mikolov2013linguistic}. These vectors, known as word embeddings, have been applied to analyze embedded semantic meanings of concepts such as equality \autocite{rodman2020timely}, class \autocite{kozlowski2019geometry}, and incivility \autocite{liang2023word} across spatial, temporal, and cultural contexts. 

In a word embedding model, each unique word appearing in a document is represented by a vector \autocite{mikolov2013linguistic,pennington2014glove} that positions it in a high-dimensional geometric space in relation to every other unique word. A word’s adjacent neighbors in the vector space are usually words with related meanings, including the word’s own syntactic variants or synonyms. The geometric relationship, or distance, between two vectors signals the semantic (dis-)connections of the corresponding words. Such distance, or the lack thereof, is commonly quantified by the cosine similarity between these two vectors. Many word embedding methods have been proposed in the past decade. Among these, Word2vec stands out as one of the most widely used. Introduced by Mikolov and colleagues in 2013, Word2vec employs a two-layer neural network to process text by vectorizing words: its input is a text corpus, and the output is a set of vectors that represent words in that corpus. In our following analyses, we demonstrate how even a plain word embedding model can be integrated with a dictionary to enhance the model performance in measuring latent moral signals from texts. 

Words that are geometrically clustered can indicate a latent semantic concept, constructing a representation of a latent concept is thus analogous to building a word representation in the vector space. The DDR approach \autocite{garten2018dictionaries} utilizes the average of vector representations of the words in a dictionary to represent a given concept, or an embedded message feature like moral content in our context. For instance, the \textit{care}-relevant content can be represented by computing the average of vectors associated with \textit{care}-related words like [\textit{kindness}, \textit{compassion}, \textit{nurture}, \textit{empathy}]. The DDR approach, then, facilitates the computing of a continuous similarity metric between a moral foundation and a text. This is achieved by projecting the text into the same vector space and then calculating the similarity between the vectorized text and the axis representing the moral foundation. 

Concepts such as moral foundations often entail valences, such as \textit{care} and \textit{harm} serving as the two anchors for the \textit{Care/Harm} moral foundation. Using the DDR approach, one can construct a concept representation of, for example, the virtue of \textit{care} by averaging all relevant words associated with care per se. However, the representation of the vice of \textit{harm} remains a challenge—even though one can similarly construct a \textit{arm} axis by averaging the vectors of \textit{harm}-related words, this axis usually would not be geometrically positioned as the opposite anchor to care on the same \textit{Care/Harm} axis. To address this limitation, An et al. \autocite{an2018semaxis}, also see a similar approach in \autocite{sagi2014measuring}, proposed SEMAXIS, a framework that creates an integrated vector axis for a target concept, encompassing both its positive and negative aspects (e.g., the virtue and the vice for a moral foundation) as the two opposing anchors on the shared axis. It can be understood as a “concept axis” in a vector space. Analyzing such axis allows us to measure the semantic similarity of documents, composed of individual words, relative to these concept axes.

In this context, a concept axis, or a moral axis in our case, is anchored by an antonym pair, such as \textit{Care-Harm}, \textit{Fairness-Cheating}, or \textit{Authority-Subversion}. Each antonym pair typically includes a set of the most positive (or rightness) words on one end and the most negative (or violation) words on the other \autocite{an2018semaxis}. To calculate the concept axis, for example, the \textit{Care/Harm} moral axis, the positive anchor (i.e., the virtue of \textit{care}) is first built using the DDR method by averaging the vectors of all positive words, and a similar process is applied for building the negative anchor (i.e., the vice of \textit{Harm}). SEMAXIS then finds the semantic axis that connects the negative anchor with the positive by taking the difference between the averaged vectors of two sets of pole words, i.e., the positive and negative words, related to this moral foundation (for details, see \autocite{an2018semaxis}, Equation 1). Thus, once the moral axis vector is obtained, researchers can compute the cosine similarity between a word vector and the axis to quantify the moral relevance of a single word or a text \autocite{kwak2021frameaxis}. That being said, integrating all pole words from a well-established dictionary into building moral axes comes with several challenges awaiting for solutions: words often contribute differently to a specific concept they are associated with, for instance, the word “\textit{murder}” likely contributes more to the \textit{Care/Harm} axis than “\textit{slap},” and assigning the right weight to each pole word when constructing the concept axis is both conceptually and statistically challenging. In the following sections, we will provide a detailed explanation of how our model has effectively tackled these challenges using an optimization algorithm and thus lifting the advantages of two conventional approaches into one. 

\section{The Vec-tionaries Approach and the Construction of the Moral Foundations Vec-tionary}
\label{section:vectionary_approach}

\begin{figure}[hbt!]
\centering
\includegraphics[width=0.75\linewidth]{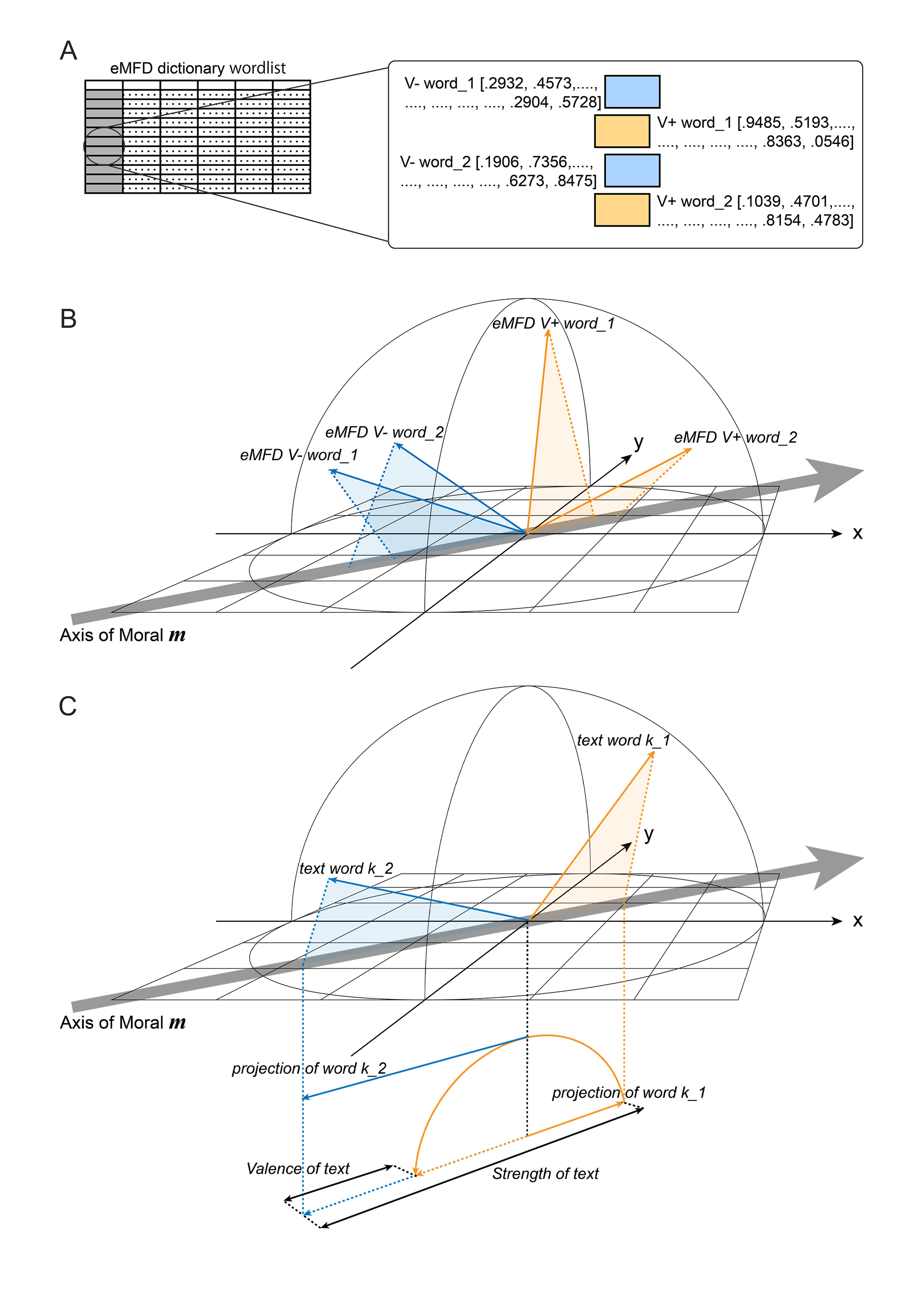}
\caption{The model pipeline of Moral Foundations Vec-tionary}
\label{fig_1}
\end{figure}

In this work, we introduce a novel framework, called vec-tionary, that integrates the well-established dictionary (i.e., eMFD) with word embedding models to measure moral content embedded in textual data. Specifically, we constructed the proposed moral foundations vec-tionary following three steps for a chosen moral foundation: 1) vectorizing words in the eMFD based on a preselected word embedding model, 2) estimating the axis for the targeted moral foundation through a nonlinear optimization algorithm, and 3) calculating the geometric distance between an unseen text and this estimated moral axis in the same vector space to derive metrics of interest (i.e., \textit{Strength}, \textit{Valence}, and \textit{Ambivalence} of a targeted moral foundation). See Figure \ref{fig_1} for an illustration of the pipeline to construct the moral foundations vec-tionary. 

We assume that the axes representing different moral foundations exist in the shared vector space with words contained in the eMFD and our goal is to uncover these axes’ geometric coordinates. We leverage eMFD’s large vocabulary and crowdsourced “weights” indicating the semantic relationships with each moral foundation. We treat each weighted word as an “observed signal” of the latent moral axis. Employing a nonlinear optimization algorithm, we iteratively update our estimates for the coordinates of the moral axes to best account for the observed weighted words from eMFD, which are themselves embedded in the same vector space. In our analyses, we used the 300-dimensional embeddings from the word2vec model, which covers nearly 3 million words and phrases. In the following sections, we present the technical details of the vec-tionary approach.

\subsection{Mathematical framework}
\label{section:vect_math}

As illustrated in Figure \ref{fig_1}, first, we transformed each eMFD word to word vectors in the semantic space. According to the assumption of the eMFD, each word $i$ is linked to all five moral foundations (except for Liberty/Oppression, a latestly added moral foundation not included in the eMFD), albeit with varying weights. The analytical goal of the moral foundations vec-tionary is to infer the coordinates for a moral axis $\boldsymbol{m}$. 

Second, we defined the \textit{observed relevance} ($s_i$) of an eMFD word as its association with a target moral foundation, already available in eMFD through a crowdsourcing procedure. Specifically, each word’s observed relevance can be obtained by merging two pieces of key information from the eMFD: the probability and the sentiment scores of the word. In the eMFD wordlist, each word was assigned a probability score (ranging from 0 to 1) for its relevance to a specific moral foundation through crowd-sourced annotations \autocite{hopp2021extended}. Additionally, the eMFD captures the sentiment valence of each MFD word per foundation, which ranges from -1(most negative sentiment, associated with moral vices) to +1 (most positive sentiment, associated with moral virtues). We merged the probability score and the sentiment valence of each eMFD word to operationalize observed relevance, where $s_i$,  $p_i$, and $v_i$ represent the observed relevance, probability score, and the valence of word $i$ in the eMFD, correspondingly. As an example, consider the eMFD word “$kill$” with a Care/Harm foundation probability score of 0.40 and a sentiment score of -0.70. We incorporated the negative valence of the sentiment score (“-1”) into the probability score, yielding an observed relevance of -0.40 for “\textit{kill}”.

While the observed relevance $s_i$ was directly obtained from human annotations during the development of the eMFD, it cannot be directly repurposed to uncover the moral axes in the vector space. To do so, we defined the \textit{analytical relevance} of an eMFD word regarding a moral foundation, denoted as $\hat{s_i}$, as the scalar projection of that word’s embedding on a particular moral axis. For example, for an eMFD word $i$ in a 300-dimension vector space, its analytical relevance, $\hat{s_i}$, represents its scalar projection onto the moral axis $\boldsymbol{m}$, where word vector $\boldsymbol{w_i}  = \left( w_{i,1}, w_{i,2},\cdots, w_{i,300}\right)$,  $\boldsymbol{m}= \left(m_1, m_2, \cdots, m_{300} \right)$. We can derive $\hat{s_i}$ as follows in Equation \eqref{eq:eq1}, where $\theta$ is the angle between vector $\boldsymbol{w_i}$ and $\boldsymbol{m}$, and $\left\lVert\cdot\right\rVert$ represents the 2-norm. To simplify the calculation, we normalized the word vector $\boldsymbol{w_i}$. Therefore, the analytical relevance $\hat{s_i}$ is essentially the cosine similarity between eMFD word vector $\boldsymbol{w_i}$ and the presumed moral axis $\boldsymbol{m}$. Cosine similarity is a standard measure in semantic vector space, which uses the cosine value of the angle between word vectors to measure their relevance \autocite{mikolov2013linguistic,an2018semaxis}. In this case, the analytical relevance $\hat{s_i}$ captures how closely the word is aligned to the moral axis $\boldsymbol{m}$.

\begin{equation}
\begin{aligned}\label{eq:eq1}
\hat{s_i} = \cos \theta \cdot \left\lVert\boldsymbol{w_i}\right\rVert = \dfrac{\boldsymbol{w_i}\cdot \boldsymbol{m}}{\left\lVert\boldsymbol{w_i}\right\rVert \cdot \left\lVert\boldsymbol{m}\right\rVert} \cdot \left\lVert\boldsymbol{w_i}\right\rVert
\end{aligned}
\end{equation}

Next, we define the error $e_i$ between the observed relevance $s_i$  and the analytical relevance $\hat{s_i}$  for a specific word $i$, as indicated by Equation \eqref{eq:eq2}. This formulation helps define an objective function for the optimization algorithm, which seeks to identify the coordinates for the moral axis $\boldsymbol{m}$ that minimizes the summation of errors for all eMFD words, as defined in Equation \eqref{eq:eq3}, where $N$ is the number of words considered. In this study, we used the nonlinear optimization solver Ipopt 0.6.5 for estimation. Interested researchers are welcome to experiment with other optimization algorithms for their domain applications. 

\begin{equation}
\begin{aligned}\label{eq:eq2}
e_i = \left(\hat{s_i} - s_i\right)^2
\end{aligned}
\end{equation}

\begin{equation}
\begin{aligned}\label{eq:eq3}
\min \displaystyle \sum^N_{i = 1} e_i
\end{aligned}
\end{equation}

Finally, to simplify calculation, we added Equation \eqref{eq:eq4} as a constraint to normalize the moral axis $\boldsymbol{m}$ and avoid obtaining the trivial zero solution.

\begin{equation}
\begin{aligned}\label{eq:eq4}
\left\lVert\boldsymbol{m}\right\rVert = 1
\end{aligned}
\end{equation}

To summarize, the proposed model includes an objective function (3) with three equality constraints defined in (1), (2), and (4). The key output is the estimated coordinates of the moral axis $\boldsymbol{m}$. The main input data includes the eMFD wordlist, along with their vector representations and observed relevance values. The pipeline is implemented in Python 3.8 (for data processing) and Julia 1.6.2 (for optimization). Specifically, JuMP 0.21.10 and Ipopt 0.6.5 are used to solve the optimization problem, which was completed within 120 seconds for a 300-dimensional vector space and a total of 3,270 eMFD words.

\subsection{Three measurement metrics}
\label{section:vect_metrics}

Compared to the dictionary approaches (e.g., eMFD), the moral foundations vec-tionary has the advantage of providing multiple metrics to capture more nuanced aspects of moral content in textual data. Beyond measuring the magnitude of moral content in a text (\textit{Strength}), the vec-tionary also captures the degree of expressed virtue versus vice for a particular moral foundation (\textit{Valence}). Additionally, our approach also measures the degree of variance among the virtue-vice moral axis for a particular type of moral content (i.e., \textit{Ambivalence}), to some degree capturing moral conflict such as the co-existence of both virtue- and vice-related expressions in a document. This Ambivalence metric is not directly available in previous moral foundations dictionaries. This expanded range of metrics not only enriches the scope of analysis for moral content but also bolsters the utility of vec-tionaries in computational analysis of message features.

Specifically, the first metric, \textit{Strength}, is denoted as the averaged \textit{absolute values} of word-level projections (i.e., cosine similarities) of a document, as indicated by Equation \eqref{eq:eq5}, where $n$ represents the number of words in a document, and $\theta_i$ is the angle between the vector representation of word $i$ and the obtained moral axis $\boldsymbol{m}$. The \textit{Strength} score ranges from 0 to 1, with larger values indicating a stronger moral foundation-specific relevance in the document regardless of valence. 

The second metric, \textit{Valence}, calculates the averaged word-level cosine similarities, ranging from -1 to 1, see Equation \eqref{eq:eq6}. It evaluates whether a document leans towards one side of a target moral foundation, with a higher \textit{Valence} score indicating the use of virtue-dominated moral expressions, and a negative \textit{Valence} score indicating vice dominance. Virtue- versus vice-related moral expressions might nullify each other. For example, if a conservative tweet may talk about “\textit{saving immigrants’ lives}” in the context of “\textit{threatening local community safety},” these two would cancel each other out. 

The last metric, \textit{Ambivalence}, calculates the variance of word-level cosine similarities, ranging from 0 to 1, as defined in Equation \eqref{eq:eq7}. This metric captures the variability in word-level moral cues in a document. A higher \textit{Ambivalence} score indicates a stronger co-presence of both virtue- and vice-related expressions. For instance, in the same example, despite the overall valence being low, the resulting high \textit{Ambivalence} score suggests appealing to both sides of the \textit{Care/Harm} moral foundation within the tweet.

%%% Numbered equation
\begin{equation}
\begin{aligned}\label{eq:eq5}
S = \dfrac{\displaystyle\sum_{i=1}^n\left|\cos \theta_i\right|}{n}
\end{aligned}
\end{equation}

\begin{equation}
\begin{aligned}\label{eq:eq6}
V = \dfrac{\displaystyle\sum_{i=1}^n\cos \theta_i}{n}
\end{aligned}
\end{equation}

\begin{equation}
\begin{aligned}\label{eq:eq7}
A = \dfrac{\displaystyle\sum_{i=1}^n\left(\cos \theta_i - V\right)^2}{n}
\end{aligned}
\end{equation}

The moral foundations vec-tionary offers several advantages over previous measurement strategies. First, it establishes moral axes based on a large, validated 3,270 dictionary words, rather than relying on a limited number of seed words (see two examples in Table A1, Supplementary Materials A), which enhances comprehensiveness and robustness. Secondly, the moral foundations vec-tionary recognizes that different words may contribute differently to a moral dimension, unlike previous that simply assumed equal seed word contributions. Lastly, constructing moral axes with varying word weights presents mathematical challenges, as the problem of uncovering coordinates in a high-dimensional vector space is non-trivial to solve. We took advantage of a non-linear optimization algorithm to extract the maximal amount of moral signals from all eMFD words along with their corresponding weights. To our best knowledge, this represents the first attempt in the literature on computational analyses of moral content to enhance an established dictionary with word embeddings and a formal optimization algorithm. Next, we present empirical evidence comparing the performance of the moral foundations vec-tionary with eMFD benchmarked on a “ground truth” tweets dataset in a context of a politicized public health crisis which diverse moral discussions have been widely merged.

\section{Model Validation and Performance Comparison}
\label{section:model_validation}

We validated the performance of moral foundations vec-tionary against the eMFD on a benchmark dataset of COVID-19 tweets annotated for the moral \textit{Strength} through a crowdsourcing procedure.

\subsection{Annotators, training, and annotation procedure}
\label{section:valid_anno}

\subsubsection{Annotation platform}
\label{section:valid_anno_plat}
We developed a crowdsourcing system that implements the pairwise comparison task built based on the open-sourced “All Our Ideas” project \autocite{salganik2015wiki}, also known as the “wiki-surveys”(www.allourideas.org), see details of our customized platform in Supplementary Materials B. For each moral foundation, we created two tasks, one measuring the virtue aspect of the foundation and the other the vice aspect (e.g., one question on \textit{care} and the other on \textit{harm} for the \textit{Care/Harm} foundation), gathering human annotators’ moral judgments on a same set of tweets, consistent with prior practices \autocite{hoover2020moral}.

The statistical rationale for the pairwise comparison task and the procedures to estimate per-message moral scores from the annotations results are detailed elsewhere \autocite{salganik2015wiki}. In a nutshell, the system constructs an opinion matrix based on respondents’ selected tweets from each pair (see an example task interface in Figure B1, Supplementary Materials B) and estimates the latent score for each tweet through Bayesian inference and a hierarchical probit model. Conceptually, the resulting latent score for a particular tweet, ranging between 0 and 100, can be interpreted as its likelihood of outperforming a randomly chosen tweet for a randomly selected annotator: a minimum of 0 indicates consistent loss while a maximum of 100 means the tweet would always win. For instance, when assessing the virtue of \textit{care}, a tweet that reads,“\textit{After ousting a dictator, members of Sudan’s resistance committees are now helping to fight the Covid-19 pandemic},” receiving a score of 96, suggests that for a random annotator, this tweet would be estimated to outperform a randomly selected tweet 96\% of the time. Finally, foundation by foundation, we were able to construct an overall ranking of all the annotated tweets based on the estimated moral \textit{Strength} scores from the crowdsourcing system.

\subsubsection{Annotators recruitment and training}
\label{section:valid_anno_annotar}

For each moral foundation, to produce sufficient data density \autocite{hopp2021extended,carlson2017pairwise}, we ensured that at least 70\% of the tweets in the stimuli pool will be evaluated by at least 15 annotators (for calculation details, refer to Supplementary Materials C). 

Annotators were recruited from the Prolific platform and each of them was assigned two tasks: one focused on the virtue dimension and the other on vice, each task involving at least 25 pairs of tweets. To avoid potential order effects, we randomized the sequence of these two annotation tasks. Furthermore, we matched this sample to census distributions on five key demographic variables: gender, age, ideological affiliation, education, and race (descriptive statistics details see Supplementary Materials D).

Each annotator focuses on one randomly assigned moral foundation. Before tasks, they are invited to an online training module, and only those who pass are eligible to proceed to annotation tasks (training materials in Supplementary Materials E). After further screening to exclude annotators who fail the test and those were timed out, we retained a total of 3,473 qualified annotators in the analytical sample, informed consent was obtained.

\subsubsection{Stimuli corpus for annotation}
\label{section:valid_anno_stim}
We collected tweets from June 15 to July 12, 2020, through Twitter’s COVID-19 firehose API. Since Twitter’s original search query includes non-English terms, we applied core 25 keywords (see Table A2, Supplementary Materials A) to further filter the corpus to make our dataset more focused. After preprocessing, this procedure resulted in a total of 2,285,379 unique English tweets. Tweets contain moral content \autocite{hoover2020moral}, while the overall prevalence could be low, thus we stratified sampled tweet stimuli by eMFD scores. To ensure sufficient variance in our stimuli corpus, for each moral foundation, we randomly selected 800 tweets from the strata with highest eMFD scores for the virtue and vice. Then, we added 400 tweets with low scores across all five foundations as control. This sampling strategy yielded 2,000 unique tweets per moral foundation (see Supplementary Materials F, G for details). 

\subsection{Performance comparison based on the rank-biased overlap method (RBO)}
\label{section:valid_rbo}
We assessed the moral foundations vec-tionary by comparing its outputs with the eMFD scores, using crowdsourced human annotations as the “ground truth.” Given that the three approaches (vec-tionary, eMFD, and human annotation) used different scales for moral relevance scores, our focus was on comparing the rankings of the 2,000 tweets per foundation by these methods. 

For the moral foundations vec-tionary, the ranking was built based on its \textit{Strength} scores. Conceptually, they reflect the relevance of moral foundation in texts and are therefore comparable to eMFD’s probability scores. Regarding the crowdsourced benchmark dataset, we built the ranking by summing up the square of both the virtue and vice scores for each tweet. To simplify the notations, hereafter we refer to the moral foundations vec-tionary as vectionary and the transformed crowdsourced scores aggregating virtue and vice as C.S. for brevity. 

We define the similarity between ranking $i$ and ranking $j$ as $R_{i,j}$. We used the rank-biased overlap method (RBO) to measure $R_{vectionary,CS}$ and $R_{eMFD,CS}$ accordingly, and then computed their difference $R_{vectionary,CS}-R_{eMFD,CS}$ (see the RBO Equation in Supplementary Materials H). RBO, first introduced by Webber and colleagues, is a continuous measure that quantifies the similarity between two ranked lists \autocite{webber2010similarity}. It has been used in many fields \autocite{ng2019mapping,urman2022matter} and has been shown to be more sensitive to the positions of overlapping items compared to other similarity measures. RBO takes into account both the depth of overlap (i.e., how many items are shared between the two lists) and the rank positions of the overlapping items (i.e., how close the overlapping items are to the top of the lists). It assigns more weights to items that are ranked higher than lower, aligning with our interest in measuring rank changes and assigning greater weight to the top of the list of stimuli tweets compared to those occurring further down. RBO provides adjustable parameters to systematically explore how similarities might change as the researcher places more weight to items at the top of the two rankings. In our case, RBO allows for a robustness check to assess the measurement performance of the moral foundations vec-tionary versus the eMFD while varying the degree to which tweets with stronger moral cues should dominate the calculation of similarities.

We calculated $R_{vectionary,CS}$ and $R_{eMFD,CS}$ with varying weights and depths, which jointly determine the RBO parameter, denoted as $p$, in the ranking comparison (see details in Table H1, Supplementary Materials H). This approach follows similar practices in existing studies; for example, Ng and Taneja (2019) employed RBO with a $p$ value of .973 to compare the website traffic across countries. Urman et al. (2022) utilized RBO with $p$ values of .98 and .8 to study the algorithmic curation of political information. To quantify estimation uncertainty for the difference between $R_{vectionary,CS}$ and $R_{eMFD,CS}$, we employed bootstrapping (resamples = 5,000, with replacement) to estimate the 95\% confidence intervals (CIs). 

Results shown that, irrespective of varying depths and weights, the moral foundations vec-tionary rankings consistently showed higher similarities with the crowdsourcing benchmark rankings than the eMFD for three moral foundations: \textit{Care/Harm}, \textit{Authority/Subversion}, and \textit{Loyalty/Betrayal}, see Figure 2. Regarding the \textit{Sanctity/Degradation} foundation, as the $depth$ parameter increased, the moral foundations vec-tionary showed a tendency to outperform the eMFD, albeit falling short of reaching the conventional threshold for statistical significance. Furthermore, these two methods did not significantly differ with regards to the \textit{Fairness/Cheating} foundation. To facilitate interpretation, in Supplementary Materials I, we provide exemplar tweets where the moral foundations vec-tionary produced more accurate results than the eMFD. 

\begin{figure}[hbt!]
\centering
\includegraphics[width=0.95\linewidth]{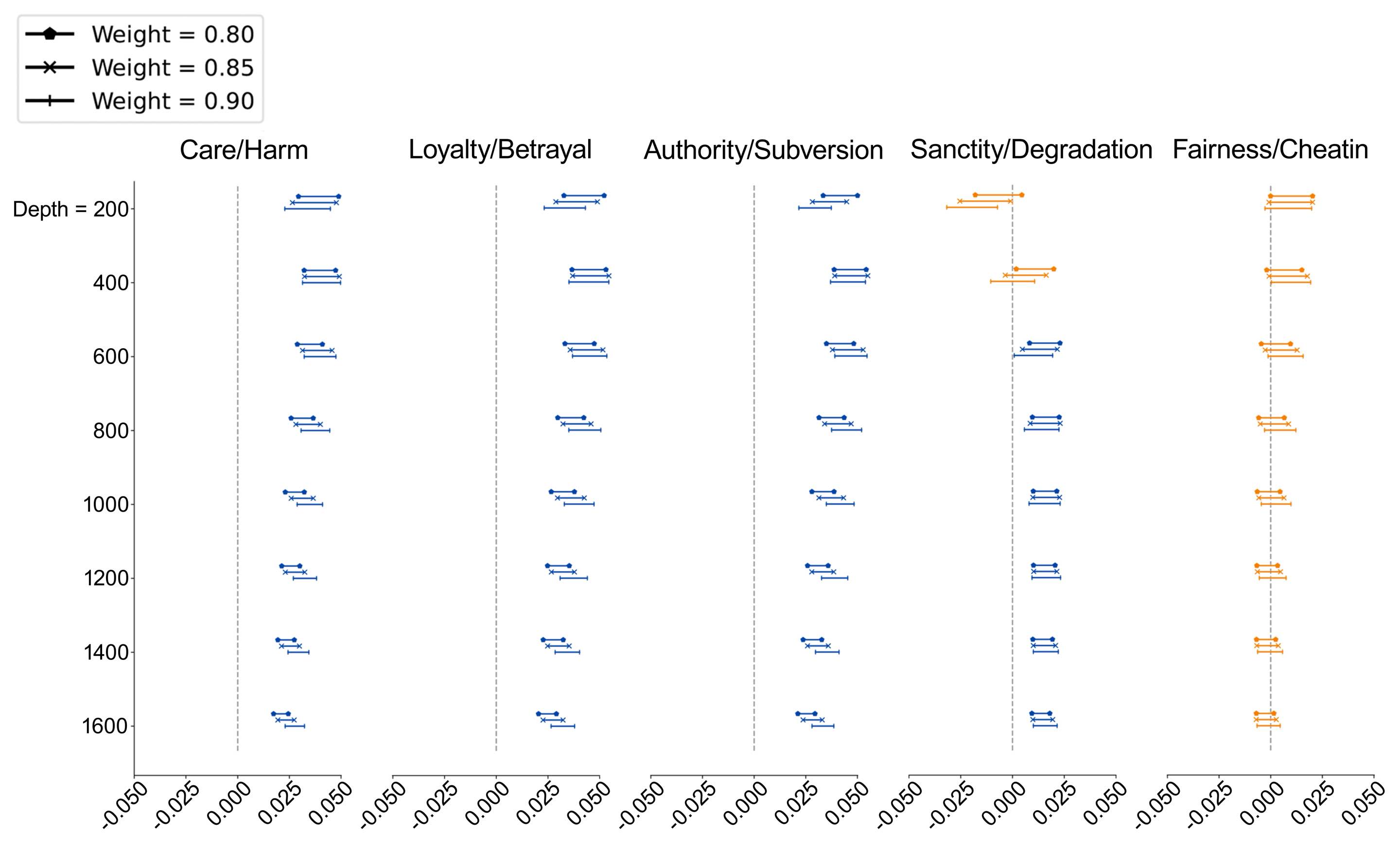}
\caption{Difference in RBO similarity scores by moral foundation}
\label{fig_2}
\end{figure}

To better quantify measurement improvement regarding the three moral foundations where the moral foundations vec-tionary outperformed the eMFD, we calculated the metric Percentage Performance Increase (PPI), see Equation \eqref{eq:eq8}. 

\begin{equation}
\begin{aligned}\label{eq:eq8}
\mathrm{PPI} = \dfrac{R_{vectionary,CS} - R_{eMFD,CS}}{R_{eMFD,CS}}
\end{aligned}
\end{equation}

We calculated the pairwise similarity scores between vectionary, eMFD, and crowdsourced benchmark, foundation by foundation, along with the PPI scores contrasting vectionary’s performance with that of eMFD. Take Table \ref{table_1} as an illustrative example, it consists of four columns. The first three columns represent comparisons among the vectionary, the C.S., and the eMFD, while the last column indicates the percentage increase of vectionary over eMFD, respectively. For instance, the first row shows that for the \textit{Care/Harm} foundation, when \textit{weight} was set as .80 and \textit{depth} as 200, the RBO similarities are as follows: $R_{vectionary,CS}=.16$, $R_{eMFD,CS}=.12$, and $R_{vectionary,eMFD}=.26$. Furthermore, a PPI score of 34.67\% indicates that the Moral Foundations Vec-tionary improves the measurement of \textit{Care/Harm} appeals by 34.67\%, compared to the eMFD, when benchmarked against crowdsourced human annotations. For the remaining foundations, we have summarized their calculations in Tables J2 to J5, which can be found in Supplementary Materials J.

\begin{table}[hbt!]
%\begin{threeparttable}
\caption{Performance Comparison for the Care/Harm Moral Foundation While Varying Weight and Depth Values}
\label{table_1}
\begin{tabular}{cccccc}

\toprule
 \headrow & & \multicolumn{3}{c}{RBO Similarities} & PPI of Vec-tionary \\
 \headrow & & Vec-tionary vs. C.S. & eMFD vs. C.S. & Vec-tionary vs. eMFD & over eMFD (\%)\\
\midrule
    & Weight = .80 & .16 & .12 &.26 & 34.67\\
    Depth 200 & Weight = .85 & .13 & .09 & .22 & 41.30 \\
    & Weight = .90 & .11 & .07 & .18 & 49.57\\
\midrule   
    & Weight = .80 & .28 & .24 &.41 & 18.07\\
    Depth 400 & Weight = .85 & .24 & .19 & .36 & 22.72 \\
    & Weight = .90 & .19 & .15 & .31 & 28.66\\
\midrule   
    & Weight = .80 & .38 & .34 &.50 & 11.26\\
    Depth 600 & Weight = .85 & .31 & .27 & .44 & 15.17 \\
    & Weight = .90 & .27 & .23 & .40 & 19.03\\
\midrule   
    & Weight = .80 & .44 & .41 &.56 & 8.38\\
    Depth 800 & Weight = .85 & .40 & .36 & .52 & 10.29 \\
    & Weight = .90 & .33 & .29 & .46 & 14.19\\
\midrule   
    & Weight = .80 & .50 & .47 &.61 & 6.55\\
    Depth 1000 & Weight = .85 & .44 & .41 & .56 & 8.38 \\
    & Weight = .90 & .38 & .34 & .50 & 11.26\\
\midrule   
    & Weight = .80 & .53 & .51 &.63 & 5.67\\
    Depth 1200 & Weight = .85 & .50 & .47 & .61 & 6.55 \\
    & Weight = .90 & .42 & .38 & .54 & 9.33\\
\midrule   
    & Weight = .80 & .57 & .55 &.66 & 4.81\\
    Depth 1400 & Weight = .85 & .53 & .51 & .63 & 5.67 \\
    & Weight = .90 & .47 & .44 & .58 & 7.45\\
\midrule   
    & Weight = .80 & .61 & .59 &.70 & 4.00\\
    Depth 1600 & Weight = .85 & .57 & .55 & .66 & 4.81 \\
    & Weight = .90 & .50 & .47 & .61 & 6.55\\
\bottomrule

\end{tabular}
\begin{tablenotes}[hang]
\item[]Table note
\item[a]Vec-tionary = moral foundations vectionary; C.S. = crowdsourced benchmark scores; PPI = percentage performance increase
\end{tablenotes}
%\end{threeparttable}
\end{table}

\begin{figure}[hbt!]
\centering
\includegraphics[width=0.75\linewidth]{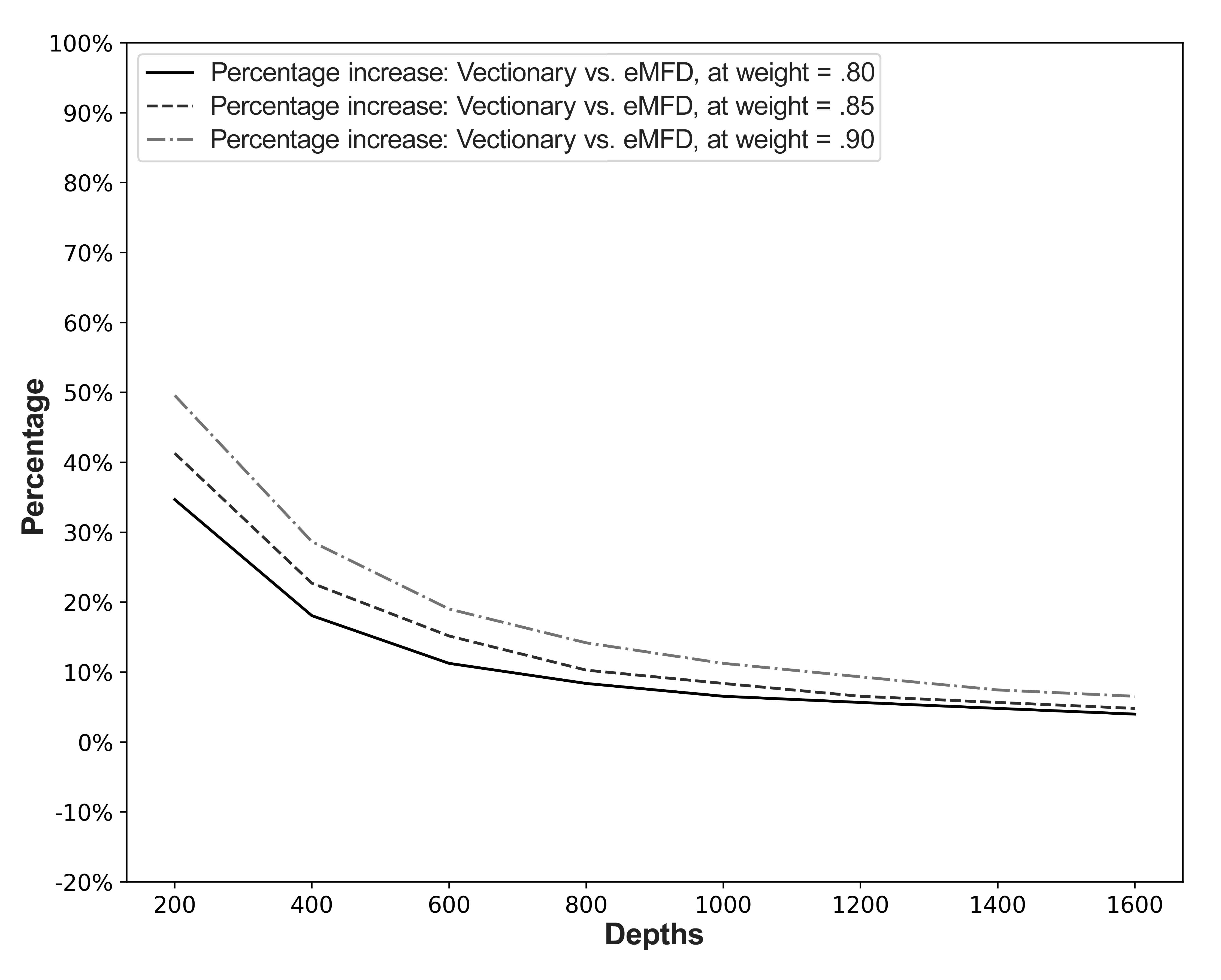}
\caption{Performance gain of the Moral Foundations Vec-tionary: Care/Harm}
\label{fig_3}
\end{figure}

\begin{figure}[hbt!]
\centering
\includegraphics[width=0.75\linewidth]{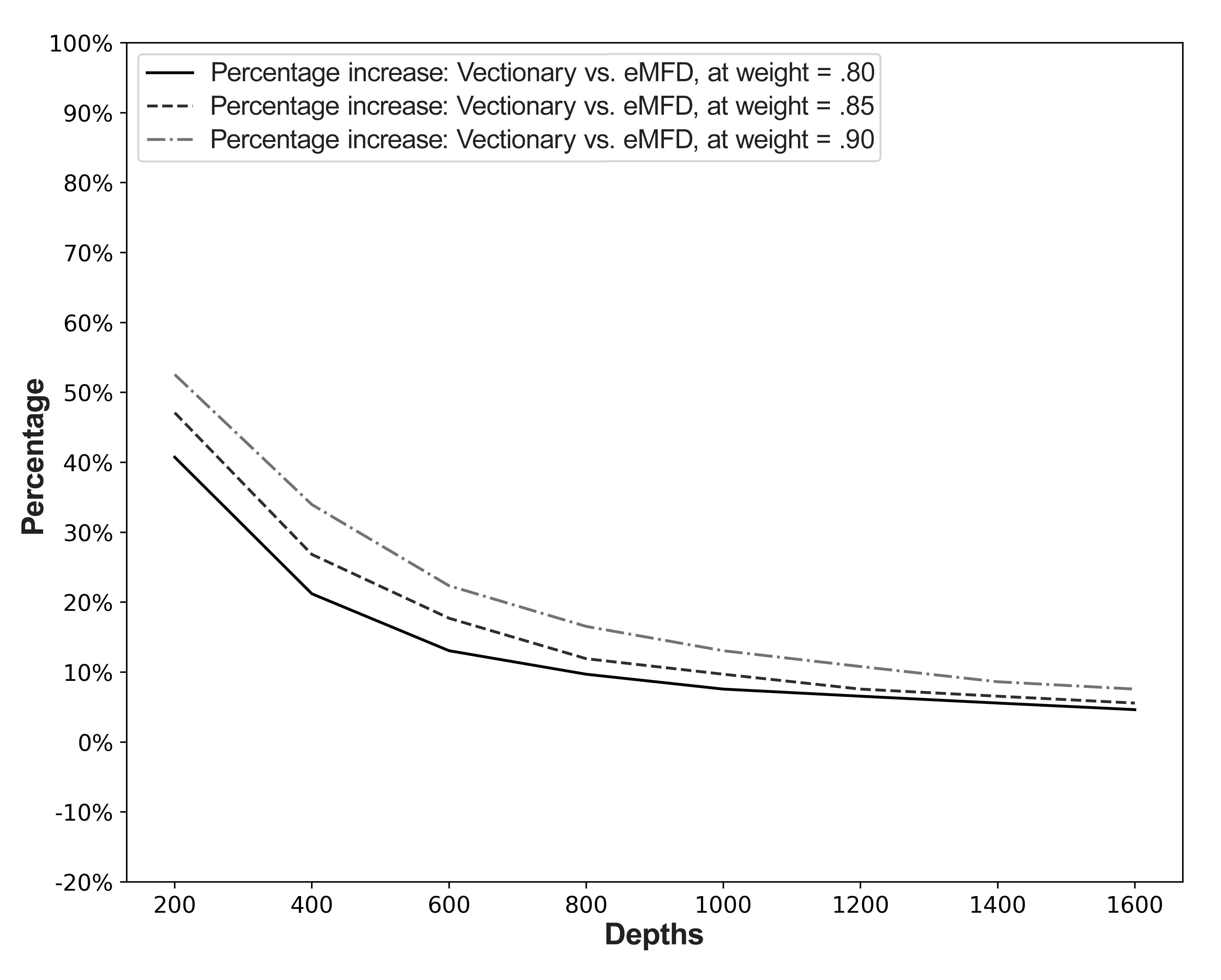}
\caption{Performance gain of the Moral Foundations Vec-tionary: Loyalty/Betrayal}
\label{fig_4}
\end{figure}

\begin{figure}[hbt!]
\centering
\includegraphics[width=0.75\linewidth]{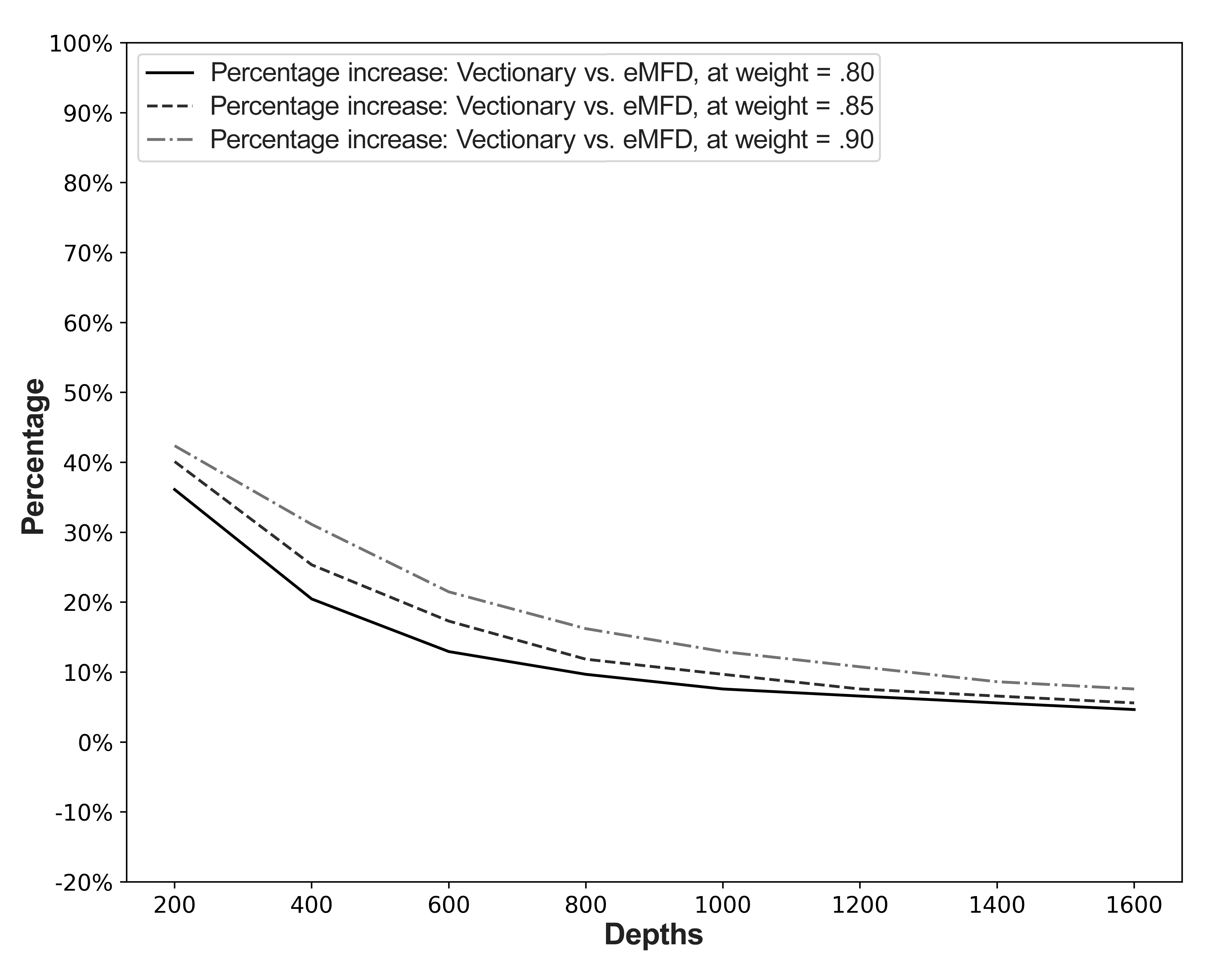}
\caption{Performance gain of the Moral Foundations Vec-tionary: Authority/Subversion}
\label{fig_5}
\end{figure}

In Figures \ref{fig_3},\ref{fig_4},\ref{fig_5}, we visualized the PPI scores for three moral foundations (\textit{Care/Harm}, \textit{Authority/Subversion}, and \textit{Loyalty/Betrayal}). The results show that the moral foundations vec-tionary tends to outperform the eMFD more with lower \textit{depth} values and higher \textit{weight} values. This pattern suggests that the moral foundations vec-tionary is particularly sensitive to capture stronger moral cues within texts because the combination of lower \textit{depth} and higher \textit{weight} would correspond to prioritizing top-ranked tweets, for a given moral foundation, in similarity calculation (see Supplementary Materials J for more details). This property of our moral foundations vec-tionary is arguably desirable, as many research applications would focus on social media posts that contain strong and clear moral signals.

\section{An Application of the Moral Foundations Vec-tionary}
\label{section:application}

Public opinion scholars have long been intrigued by the theoretical question of which specific message features, such as moral content, function as “triggers” for increased online retransmission \autocite{brady2017emotion,brady2019ideological}. In this study, we aim to illustrate how the moral foundations vec-tionary can effectively identify moral content within tweets. Additionally, we explored its ability to predict the number of retweets, surpassing the eMFD scores, after controlling for common covariates. Furthermore, we sought to assess whether additional measurement metrics, namely, moral \textit{Valence} and \textit{Ambivalence}, which are not available in the eMFD, can account for unique variances beyond moral \textit{Strength}.

Prior to fitting the models, we applied standard text-preprocessing procedures to the corpus of COVID-19 tweets (details available in Supplementary Materials G). Given our interest in predicting the number of retweets as a case study to illustrate the usefulness of measures from the moral foundations vec-tionary, starting from June 15, we guaranteed that each tweet in our corpus had an equal chance to accrue retweets by applying an identical 14-day moving window. Additionally, we incorporated metadata such as account verification status and expressed emotion valence as control variables. The main outcome, the number of retweets, is a count variable with skewed distribution characterized by over-dispersion and a high proportion of zeroes (78.69\% of the total dataset). Therefore, we employed the Zero-Inflated Negative Binomial (ZINB) regression to examine the relationships between moral content and retweeting.

\begin{figure}[hbt!]
\centering
\includegraphics[width=0.75\linewidth]{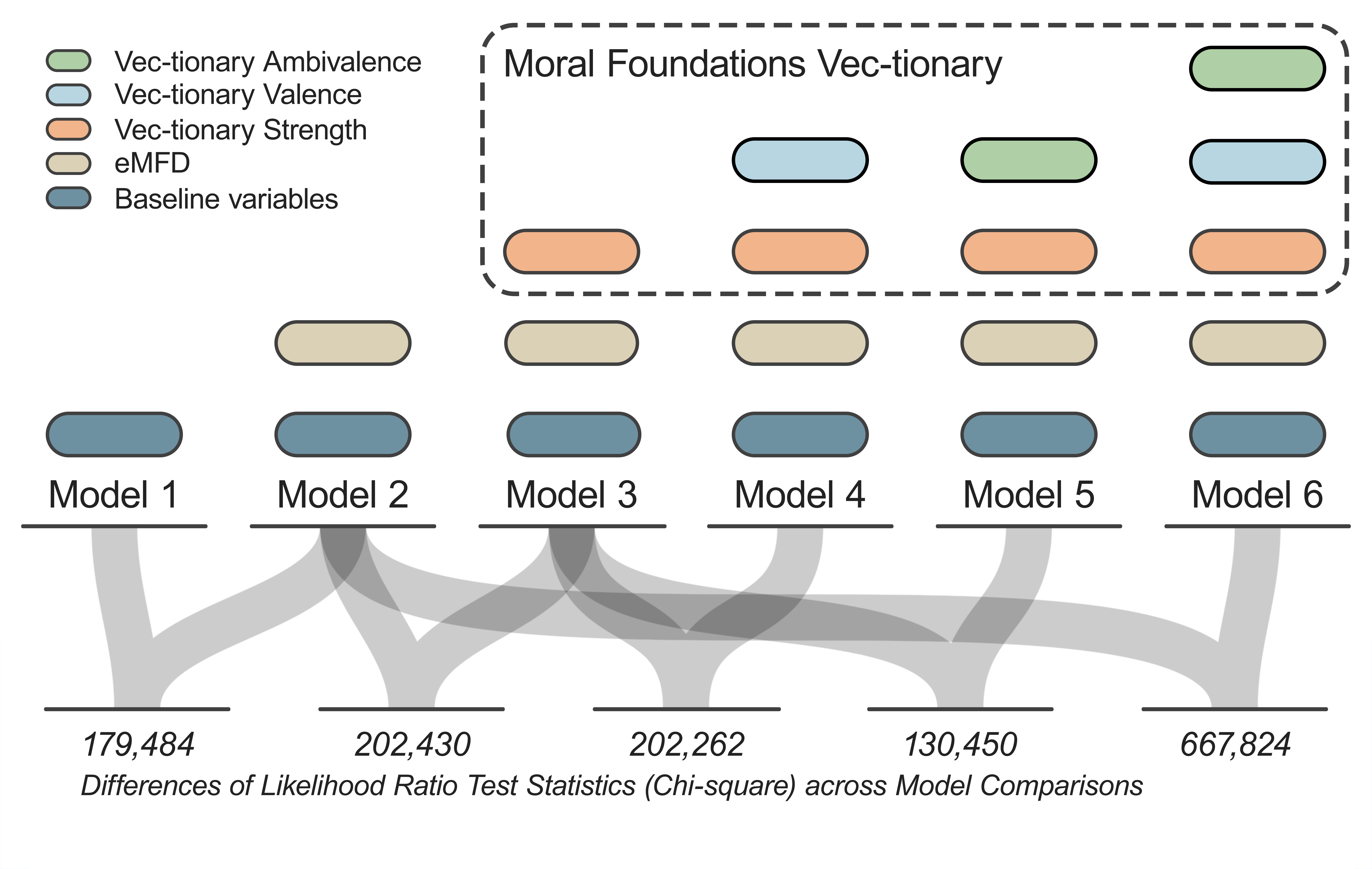}
\caption{Model specification and comparison}
\label{fig_6}
\end{figure}

Figure \ref{fig_6} summarizes model specification for the six models that we fit to assess the predictive power of moral content: Model 1 was the baseline model with only meta data and expressed emotions; Model 2 added eMFD scores to Model 1, and Model 3 in turn added moral \textit{Strength} scores from the moral foundations vec-tionary to Model 2; Model 4 and 5 respectively added moral \textit{Valence} and \textit{Ambivalence} scores to Model 3; and Model 6 was the “kitchen sink” full model incorporating all predictors previously mentioned. The addictive structure of these models allows us to unpack whether metrics from the moral foundations vec-tionary can account for unique variances in predicting the number of retweets, through a series of likelihood ratio tests (LRTs) \autocite{lewis2011unified}. We also assessed changes in Akaike Information Criterion (AIC) and Bayesian Information Criterion (BIC) as complementary evidence. To address potential multicollinearity concerns, we assessed the Variance Inflation Factor (VIF) for each model and observed VIF values within the acceptable range of 1.01 to 9.40, indicating no significant multicollinearity issues given the context of our large dataset and complex models.

Supplementary Materials K (Table K4) presents model comparison results. Echoing prior research \autocite{brady2017emotion}, Model 2 incorporating eMFD probability scores significantly improved model fit over the baseline Model 1 ($\Chi^2(10) = 179,484; \Delta_{AIC} = -179,465, \Delta_{BIC} = - 179,338$), demonstrating that moral content measured through eMFD significantly predicted retweeting. Model 3, adding moral \textit{Strength} measures from the moral foundations vec-tionary, explained additional variances than Model 2 ($\Chi^2 (10) = 202,430;  \Delta_{AIC}= - 202,409, \Delta_{BIC}= -202,283$), suggesting that the moral foundations vec-tionary captured unique moral signals beyond eMFD.

Next, we assessed whether the two additional metrics from the moral foundations vec-tionary, i.e., moral \textit{Valence} and \textit{Ambivalence}, enhanced predictive power beyond moral \textit{Strength} and the eMFD measures. Model 4 and 5, adding \textit{Valence} and \textit{Ambivalence} scores, respectively, both outperformed Model 3 ($\Chi^2(10) = 202,262 $ for Model 4; $\Delta_{AIC} = - 202,241, \Delta_{BIC} = -202,115; \Chi^2(10) = 130,450,$ for Model 5; $\Delta_{AIC} = -130,431, \Delta_{BIC}= -130,305$). Lastly, comparing the full model (Model 6) with Model 2, we found a significant model fit improvement by incorporating all three metrics from the moral foundations vec-tionary ($\Chi^2(30) = 667, 824; \Delta_{AIC}=- 667,764, \Delta_{BIC}= - 667,385$).

In summary, our findings consistently demonstrate improved model performance when incorporating the three metrics from the moral foundations vec-tionary. Though conceptually similar to the eMFD moral scores, the moral \textit{Strength} metric from the moral foundations vec-tionary accounted for unique variances in predicting retweeting beyond the eMFD. Furthermore, the two additional metrics, moral \textit{Valence} and \textit{Ambivalence}, offered unique explanatory power. Therefore, researchers can benefit from adopting the three distinct metrics that the moral foundations vec-tionary provides for a multifaceted assessment of in-text moral content.

\section{Discussion}
\label{section:discussion}
We introduce a novel computational approach, named vec-tionaries, to extract and measure message features from texts. In this paper, we focus on moral content as a case study, due to growing scholarly interest in studying the roles of moral content in public opinion, political engagement and persuasion, online communicative behaviors, among others \autocite{feinberg2013moral,graham2013moral}. The moral foundations vec-tionary draws from an extensive methodological literature on measuring moral content, notably the eMFD based on crowdsourcing \autocite{hopp2021extended} and the DDR method based on word embeddings \autocite{garten2018dictionaries,an2018semaxis}. In constructing the moral foundations vec-tionary, we employed nonlinear optimization algorithms to estimate moral axes in a semantic vector space, by merging crowdsourced moral ratings from the eMFD with established word embeddings. 

The moral foundations vec-tionary stands out in several ways. First, its architectural framework allows outputting an array of metrics, including \textit{Strength}, \textit{Valence}, and \textit{Ambivalence}, to quantify distinctive aspects of moral content—a noteworthy expansion broadening the scope of available measures from existing various moral foundations dictionaries (i.e., eMFD), as detailed in the Section \ref{section:vect_metrics}. Moral \textit{Strength} captures the presence and magnitude of a particular type of moral content in a text, collapsing the virtue and vice dimensions of the corresponding moral foundation. Our validation analyses through the RBO analyses, refer to the Section \ref{section:valid_rbo}, have largely confirmed the superiority of the moral \textit{Strength} measure from the moral foundations vec-tionary, benchmarked against crowdsourced human annotations. Furthermore, the \textit{Valence} measure assesses the predominant moral sentiment by taking the net difference between expressed virtue and vice for a given moral foundation, whereas \textit{Ambivalence} measures the variance along the foundation-specific virtue-vice axis—for example, higher values of \textit{Ambivalence} could be interpreted as indicating higher moral conflict, i.e., mentioning both virtue and vice. In the reported application in Section \ref{section:application} predicting tweet retransmission, we not only reaffirmed our previous validation results by showing the unique variances accounted for by moral \textit{Strength} scores, but also underscored the significance of incorporating \textit{Valence} and \textit{Ambivalence}—these results remain valid even after controlling for eMFD scores, expressed emotions, and other baseline predictors. Taken together, the moral foundations vec-tionary not only yields better measurements for moral \textit{Strength}, but also opens new avenues for researchers to explore, particularly regarding moral ambiguity and conflict (through the \textit{Ambivalence} metric), where discussions about virtue and vice often co-occur within the same message. 

Another notable advantage of vec-tionaries is that, unlike traditional dictionary-based methods that consider only a limited set of keywords, vec-tionaries encompass all available words within a given text. This distinction is essential because conventional dictionaries often risk invoking false negative errors—incorrectly indicating the absence of a moral foundation—when context-specific moral signals are contained in words absent from the dictionary’s word list. In contrast, vec-tionaries employs nonlinear optimization to harness continuous ratings from the full list of eMFD words while incorporating additional moral signals from other words of a given text beyond the eMFD list. In the context of studying moral content, this property becomes especially valuable when researchers are interested in analyzing moral content within short-form texts such as social media posts \autocite{brady2017emotion,zhou2022moral}, where signals are scarce. Directly applying the eMFD to short-form social media posts might not yield accurate measurements, because the eMFD was originally developed for measuring long-form texts such as news stories. The original authors of the eMFD have also emphasized this limitation \autocite{hopp2021extended}. 

The last notable advantage of vec-tionaries is contextual adaptability captured through word embeddings. Conventional dictionary-based methods often neglect context-specific nuances. In contrast, vec-tionaries allow the selection of word embeddings tailored to specific contexts. For instance, researchers can substitute the default general-purpose word embeddings (e.g., word2vec, GloVe) with embeddings tailored to the specific context or application. The model can also incorporate word embeddings from fine-tuned large language models such as Generative Pre-trained Transformers (GPTs) as they become available.	

This study emphasizes the importance of validation by benchmarking crowdsourced data. We developed a protocol to crowdsource human annotations of moral content within short-form texts, taking insights from the pairwise comparison paradigm \autocite{carlson2017pairwise,salganik2015wiki}. Given the documented difficulty in measuring moral content following conventional manual coding procedures \autocite{weber2021extracting}, our crowdsourcing protocol fills a critical gap in the literature and can be used to construct “ground truth” datasets for moral content in other applications. Our validation results confirmed better performance of the moral foundations vec-tionary for three (i.e., Care/Harm, Authority/Subversion, and Loyalty/Betrayal) out of the five moral foundations tested, particularly for tweets containing stronger moral signals. For the remaining two moral foundations, the measurement accuracy of the moral foundations vec-tionary was on par with the eMFD. We encourage future research to replicate this documented between-foundation variation in performance in other contexts and to further investigate underlying mechanisms. Taken together, these results suggest that the moral foundations vec-tionary is a valid tool for measuring moral content from texts. That said, we do not suggest that the moral foundations vec-tionary should replace the eMFD, rather, researchers are welcome to use and test this new tool as a complementary resource to existing methods. 

Through demonstrating the validity and utility of the moral foundations vec-tionary, our aspiration is to illustrate the conceptual basis and methodological procedures for researchers interested in extending the vec-tionaries approach to measure other latent message features (e.g., emotions, incivility, linguistic sophistication, politicizing frames) across languages and contexts. Three key steps to construct vec-tionaries are worth bearing in mind: first, find a validated dictionary with wordlists and weights measuring the targeted latent feature; second, select a set of word embeddings, either general-purpose or context-specific; and finally, specify an appropriate optimization algorithm to construct semantic axes aligned with the desired latent feature(s). By following these steps, the constructed vec-tionary can yield continuous measurements for the targeted message features. In closing, we reiterate the importance of adopting an agnostic approach and conducting validation tests before using the constructed vec-tionary for substantive analyses \autocite{grimmer2022text}.

\begin{acknowledgement}
We thank Christopher Lucas, Jiaxin Pei, and Qijia Ye for their valuable insights and feedback. Additionally, we are grateful to Micol Federica Tresoldi, Chenghui Li, Ye Wang, and Doug Hemken for their statistical consulting support, and Luyu Xu for assistance with data visualization. 
\end{acknowledgement}

\paragraph{Funding Statement}
Our project receives support from the Wisconsin Alumni Research Foundation (WARF) to S.Y. (MSN231886) and K.C. (AAH2162). We also received a WARF Accelerator Big Data Challenge Grant from the same funding agency awarded to S.Y. and Z.D. (MSN275569).

\paragraph{Data Availability Statement}

The Replication Codes for this study is available in our Open Science Framework (OSF) repository (Link: \url{https://osf.io/f2bt4/?view_only=8fdb67857fc743a09eddbbc44f3ef511}).

\paragraph{Competing Interests}

None.

%\endnote in some journals will behave like \footnote; and \printendnotes will not output anything. 
\printendnotes
The Supplementary Materials for this study is accessible in our Open Science Framework (OSF) repository (Link: \url{https://osf.io/f2bt4/?view_only=8fdb67857fc743a09eddbbc44f3ef511}).

\printbibliography

\end{document}